\documentclass[format=sigconf]{acmart}
\usepackage[utf8]{inputenc}

\usepackage{amsmath}
\usepackage{algorithm}
\usepackage{algpseudocode}
\usepackage{caption}
\usepackage{dblfloatfix}

\usepackage{xcolor} % comment out to remove all colour markings
\graphicspath{{Figures/}}

\acmConference[GECCO '23]{The Genetic and Evolutionary Computation Conference 2023}{July 15--19, 2023}{Lisbon, Portugal}

\acmPrice{15.00}
\acmISBN{978-x-xxxx-xxxx-x/YY/MM} % To be updated after completing copyright process

\begin{document}
\title{Creative Discovery using QD Search}
%\titlenote{Produces the permission block, and copyright information}
%\subtitle{Subtitle}
%\subtitlenote{The full version of the author's guide is available as
 % \texttt{acmart.pdf} document}

%%% The submitted version for review should be ANONYMOUS
\author{Jon McCormack}
%\authornote{}
\affiliation{%
  \institution{SensiLab - Monash University}
  \streetaddress{900 Dandenong Road}
  \city{Caulfield East} 
  \state{Victoria}
  \country{Australia}
  \postcode{3145}
}
\email{jon.mccormack@monash.edu}

\author{Camilo Cruz Gambardella}
%\authornote{}
\orcid{1234-5678-9012}
\affiliation{%
  \institution{MADA - Monash University}
  \streetaddress{900 Dandenong Road}
  \city{Caulfield East} 
  \state{Victoria}
  \country{Australia}
  \postcode{3145}
}
\email{camilo.cruzgambardella@monash.edu}

\author{Stephen James Krol}
%\authornote{}
\affiliation{%
  \institution{SensiLab - Monash University}
  \streetaddress{900 Dandenong Road}
  \city{Caulfield East} 
  \state{Victoria}
  \country{Australia}
  \postcode{3145}
}
\email{stephen.krol@monash.edu}

\renewcommand{\shortauthors}{J. McCormack et al.}

\begin{abstract}
In creative design, where aesthetics play a crucial role in determining the quality of outcomes, there are often multiple worthwhile possibilities, rather than a single ``best'' design. This challenge is compounded in the use of computational generative systems, where the sheer number of potential outcomes can be overwhelming. This paper introduces a method that combines evolutionary optimisation with AI-based image classification to perform quality-diversity search, allowing for the creative exploration of complex design spaces. The process begins by randomly sampling the genotype space, followed by mapping the generated phenotypes to a reduced representation of the solution space, as well as evaluating them based on their visual characteristics. This results in an elite group of diverse outcomes that span the solution space. The elite is then progressively updated via sampling and simple mutation. We tested our method on a generative system that produces abstract drawings. The results demonstrate that the system can effectively evolve populations of phenotypes with high aesthetic value and greater visual diversity compared to traditional optimisation-focused evolutionary approaches.
\end{abstract}


\maketitle

\section{Introduction}

A challenge in evolutionary design is the fruitful exploration of the design space, particularly when the size of that space is vast. Traditional evolutionary models tend to focus exclusively on optimisation -- trying to find the single best design in a population according to some fitness criteria. A challenge is to formalise this fitness function so that captures the desired attributes for a specific design task. In creative disciplines, the definition of a design often emerges as the creative process unfolds due to the role of cultural and aesthetic considerations. Additionally, optimising towards a single objective can preclude the possibility of considering multiple viable designs. Combining search and optimisation capabilities with the diversity of complex generative systems can overcome this challenge.

In this paper we present a computational design system that uses quality-diversity search (QDS) \cite{lehman2008exploiting,mouret2015illuminating}, an evolutionary approach that shifts the focus from the optimisation of a singular outcome (``survival of the fittest'') to the generation of physically or visually diverse possibilities, each maximising fitness, in a similar fashion to how a designer would test alternatives in search of good designs\footnote{What constitutes a ``good'' design is complex and often subjective, For now we assume this is part of any designer's implicit knowledge.}. 

Our system operates by alternating between generation and selection. For generation we use an agent-based system, designed to produce abstract line drawings derived from the paths described by autonomous moving agents. For selection, in addition to a traditional fitness measure derived from an individual artist's aesthetic preferences, we also keep a record of the physical diversity of the alternatives in a population of drawings. This enables us to visualise and understand  the diversity of the design space in a way that preserves the most fit individuals, with the goal of maximising the number of high quality, but different, designs.

%
%==============================================================================
%
\section{A QDS-based system for creative work}
\label{s:method}

\begin{figure}
\centering
\includegraphics[width=0.7\columnwidth]{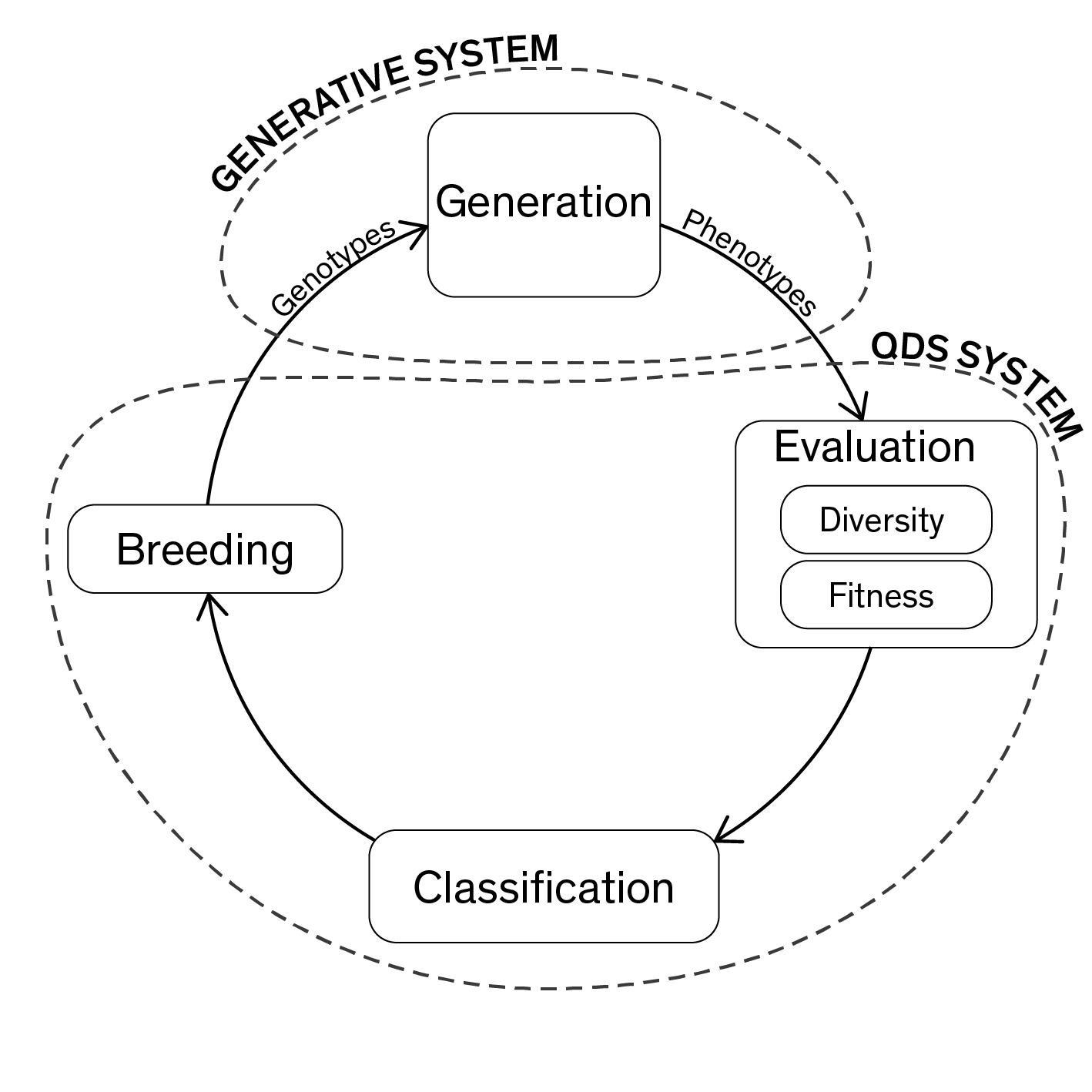}
    \caption{Four essential stages in the proposed design system: generation, two-stage evaluation, classification and breeding}
    \label{fig:method}
\end{figure}

The purpose of this study is to understand what QDS is capable of contributing in support of creative practice. The implementation of our system is built on four key processes: \emph{generation, evaluation, classification} and \emph{breeding}. The distinctive features of QDS are the evaluation of both diversity and fitness, and the classification process derived from it. These processes are executed by two interconnected systems: a generative system and an evolutionary system, illustrated in Fig. \ref{fig:method}.

Our method is intended to be used with any generative system for visual art or design, and is based on the assumption that the generative system generates artefacts from some form of parametric specification. Typically the size of the specification will be many orders of magnitude less than that of the artefacts produced \cite{Smith1984}. We refer to the parameters as the \emph{geneotype} and the generated artefact as the \emph{phenotype} (Fig. \ref{fig:method}.) We assume the following definitions:
\begin{itemize}
    \item There exists a generative system, $S$, that generates creative artefacts (2D images or 3D models, for example);
    \item $S$ takes a \textit{genome}, $G = \langle g_0,\ldots,g_n \rangle \in \mathbb{R}^n$, a vector of individual \textit{genes}, $g_i\ \in \mathbb{R}$, that determine the behaviour of $S$;
    \item $S$ produces a \textit{phenotype}, $P\ \in \mathbb{R}^d$ from $G$, i.e. $G \xrightarrow{S} P$;
    \item $P$ can be transformed to a discrete 2D image, i.e.~$P \rightarrow I = \langle r_0, g_0, b_0, \ldots r_m, g_m, b_m \rangle \in \mathbb{I}^{3m}$, where $m = x y \in \mathbb{I}$ is the total number of pixels in the image of width $x$ and height $y$ (3D models can be rendered as a 2D image from a particular viewpoint);
    \item The system designer has the ability to determine the fitness of $I$, either via observation, or using computational calculations.
\end{itemize}

The evaluation component of our system relies on being able to compute both the quality (fitness) and diversity (visual differences) of generated phenotypes. We detail how these properties are computed in the following sections.

% end of mapping explanation
\section{Experiments and Results}
\label{s:experiments}

In order to investigate how diversity contributes to a better exploration of a creative space, we used a creative generative system, designed to produce abstract line drawings. This generative system uses an agent-based line drawing model. Similar models have been effectively used in previous research, in addition to being recognised as artistically successful \cite{Annunziato1998,Baker1994,McCormack2022a}.

\begin{figure}
    \centering
    \includegraphics[width=0.5\textwidth]{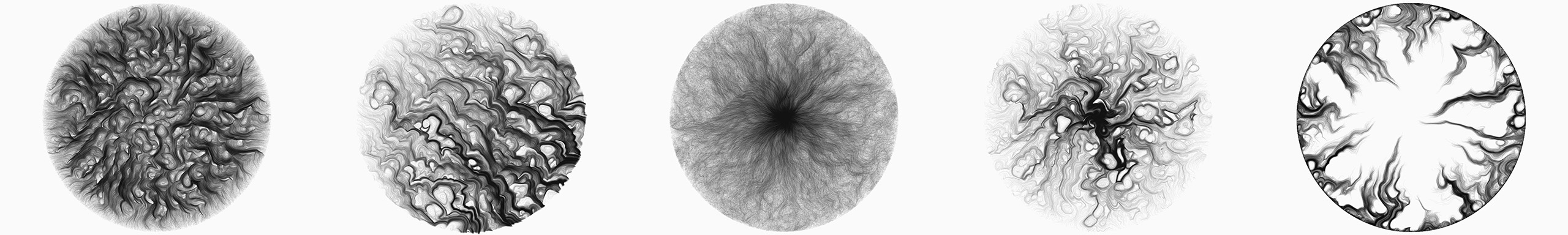}
    \caption{Example phenotypes from the agent-based line drawing system. }
    \label{fig:curlNoiseExamples}
\end{figure}

Each agent is a mobile particle that leaves an ink trace as it moves over a virtual ``canvas''. The drawing is complete when all the particles have exceeded their genetically determined lifetimes. The motion of the particles is determined by internal dynamics and flow fields based on procedural fluid flow \cite{Bridson2007}. The cumulative paths of all the particles, representing the finished drawing, are output as a \textit{png} file. The overall behaviour of each particle -- and hence the visual characteristics of the drawing they produce -- is determined by a series of 14 genetic parameters normalised to the range $[0,1]$. Together, these parameters completely determine the visual characteristics of the generated image. A selection of examples from random genotypes is shown in Fig.~ \ref{fig:curlNoiseExamples}.

\subsection{Fitness Function}
\label{ss:fitness-function}

Fitness equates to the aesthetic quality of each generated drawing. The definition of a fitness function is a complex task for ``subjective'' criteria such as aesthetics. Despite many attempts to quantify universal measures of aesthetics \cite{Johnson2019}, research from psychology \cite{Leder2014} and neuroscience \cite{Oleynick2014} shows that making an aesthetic judgement depends on many factors beyond the artefact itself. These include prior knowledge and experience, the viewing context and the affective state of the person making the judgement. Hence, developing formalised measures of aesthetics for evolutionary computing remains an ``open problem'' \cite{McCormack2005a}. One approach is to analyse an individual designer's preferences for correlation with easily computable image metrics applied to a large number (\textasciitilde{}200-1000) of random phenotypes from the generative system \cite{McCormackLomas2020b, McCormack2022a}.

For this experiment we conducted a study to determine the best correlation between a set of 14 image metrics described in \cite{McCormackLomas2020b} with the creator of the generative system's aesthetic preference. We used randomly initialised genotypes to generate 255 distinct images and asked the designer to rank these images in order of aesthetic preference. The ranking was performed using a custom-designed tool that allows the user to quickly reorder images while always showing the full list as thumbnails. The results of the study showed the highest correlation between aesthetic preference and structural complexity \cite{Lakhal2020} ($r=0.72, p < 10^{-2}$), computed by measuring the algorithmic complexity of a given image after applying a low-pass filter and tri-level thresholding to the image. Algorithmic complexity is approximated as the ratio of compressed to uncompressed versions of the image using the DEFLATE lossless-compression algorithm \cite{Deutsch1996}. The structural complexity measure is easy and fast to compute and provided a reasonable proxy to the designer's personal aesthetic preferences.

\subsection{Feature Extraction for Diversity}
\label{ss:diversity-training}

Diversity is a measure of the visual differences between artefacts. 
To assess the diversity of our evolving population of drawings, a Variational-Autoencoder (VAE) was trained on 40,000 designs generated by the line drawing system. Passing a phenotype into the trained encoder generates a feature vector that places the phenotype in a regularised latent space - this space is then used to segment different designs. The VAE feature vectors are high-dimensional (512 dimensions) so we further reduced this space using a combination of PCA to 50-D and then TSNE to 2-D. We found this helped improve elite association as our algorithm utilised the euclidean distance which is known to work differently in high dimensions. We also normalised the dimensions of the reduced vectors between 0 and 1 to ensure clusters were well distributed over both features. Once the latent space was created from the original training set, k-means was used to find initial prototypical design families in the space. The specific architecture of the VAE can be found on this Github repository \footnote{https://github.com/SensiLab/CreativeDiscoveryVAE}.

\subsection{QDS using MAP-Elites}
\label{ss:map-elites}

To perform the QD search we use a custom implementation of MAP-Elites \cite{mouret2015illuminating}, partitioning the design-solution space into regions, each of which represents a distinct type of solution. These regions are then iteratively populated with high-quality alternatives \cite{Keller2020} using a traditional fitness function. In our implementation, instead of subdividing the design space into discrete cells using a grid, or into distinct regions using Voronoi partitioning \cite{Hagg2021}, we cluster design alternatives in the design space using K-Means, to then classify solutions based on their proximity to the clusters' centroids. 

The genotypes for the individuals in the initial population $\lambda$ are randomly generated and corresponding phenotypes produced using the generative line-drawing system. Each individual in the population is then evaluated for diversity (i.e.~it is assigned to one of the $k$ pre-defined categories). If the individual's category is empty (there are no individuals from a previous population in that category) the individual occupies that category in the `elite'. Otherwise, a fitness evaluation is performed on the individual from the current population and then it is compared to the fitness of the individual already in its category in the elite. The individual with the highest fitness will occupy the category. To keep track of the performance of the evolutionary system, the fitness of a population is calculated as the average fitness of the elite, where each individual is evaluated using the fitness function described in section \ref{ss:fitness-function}. Diversity is defined as the ratio of populated clusters over total number of expected clusters. After every generation, cluster centres are updated and move towards their current elite. In early experiments, we found that the fittest individuals were rarely close to one cluster and were often between multiple clusters. By updating cluster centres, less clusters were assigned similar designs resulting in better exploration of the latent space.

In subsequent generations (i.e. once the system has already completed one generation), the genotypes for new populations are generated by sampling the elite. If an empty category is sampled, a random population is generated. Otherwise, the new population is generated by mutating the elite individual found in the sampled category using a simple mutation operator with a mutation rate $r$ that determines the probability of mutation for every allele in a gene, and a mutation factor $f$ which defines the maximum variation of a mutated allele. Below is a formulation of the algorithm:

\begin{enumerate}
    \item Create original latent space using a dataset, VAE and dimensionality reduction.
    \item Fit k-means on the latent space for \(k\) clusters to find initial prototypical designs.
    \item Generate population of $\lambda$ designs.
    \item Extract features of population using VAE and dimensionality reduction.
    \item Assign phenotypes to their closest clusters in latent space.
    \item For each cluster, assign the fittest phenotype as its' elite and update  the cluster centre closer to this elite.
    \item Repeat steps 3-6 for $e$ generations.
\end{enumerate}

\subsection{Experimental setup and results}
\label{ss:experimental-setup}

To determine values for the number of clusters $(k)$, number of generations $e$ and population size $\lambda$ that ensured quality and diversity at a low computational cost, we conducted an initial hyper-parameter search. We found $k = 25, e = 50$ and $\lambda = 15$ to be adequate for testing the system.
For the evolutionary system's inner parameters we set mutation rate to $r = 1/g_l$, where the total number of alleles in a gene $g_l = 14$, and mutation factor to $f = 0.25$.

\begin{figure}[t!]
    \centering
    \includegraphics[width=0.95\columnwidth]{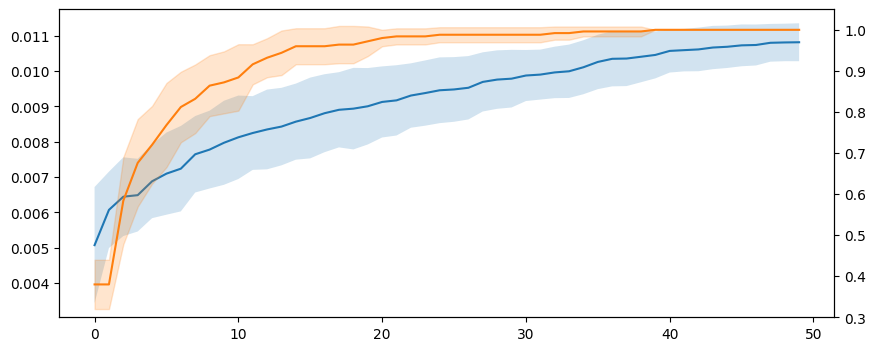}
    \caption{Aggregate time series of ten MAP-Elites runs. The blue line and shaded area (values on left $y$ axis) show the mean and standard deviation of the elite fitness. Mean and standard deviation for diversity are shown by the orange line and shaded area respectively (values on right y axis).}
    \label{fig:mapelites-aggregate}
\end{figure}

Fig.~\ref{fig:mapelites-aggregate} illustrates the aggregate results of ten MAP-elites runs, displaying consistent growth in both fitness (blue) and diversity (orange). Fitness increases gradually, while diversity exhibits a logarithmic pattern, reaching its peak between generations $e = 30$ and $e = 40$.

\begin{figure*}
    \centering
    \begin{tabular}{c|c|c}
        \includegraphics[width=.31\textwidth]{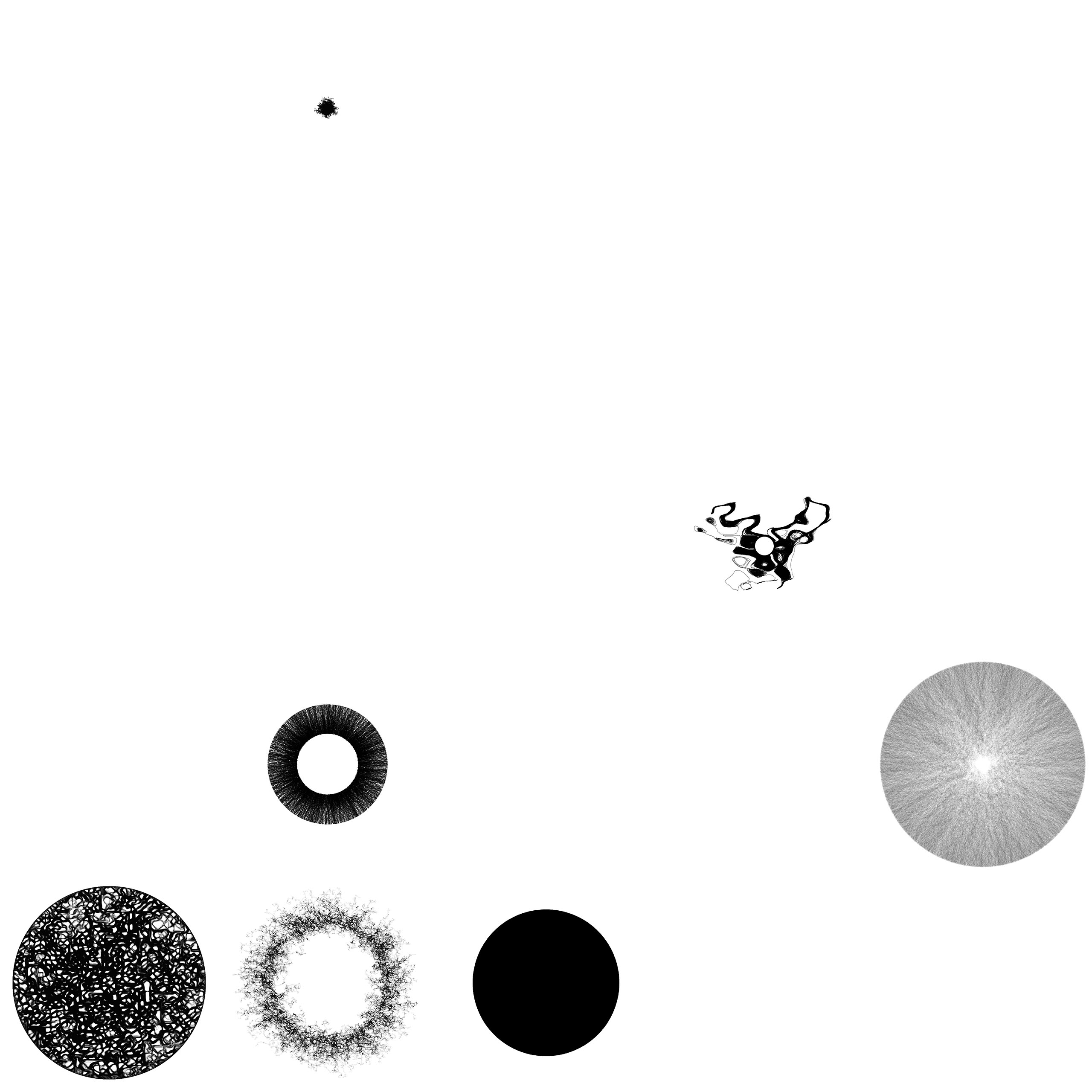} &
        \includegraphics[width=.31\textwidth]{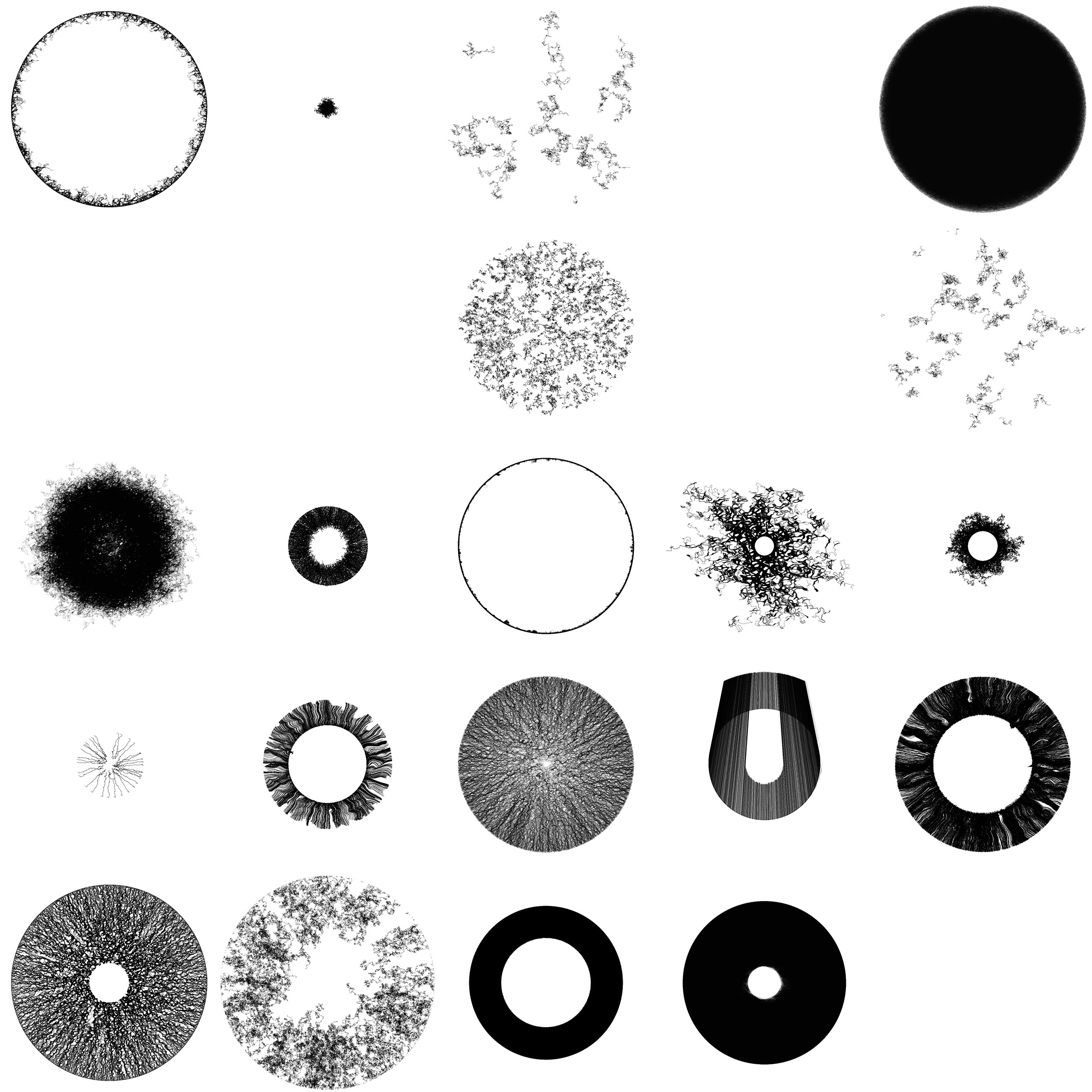}
        &
        \includegraphics[width=.31\textwidth]{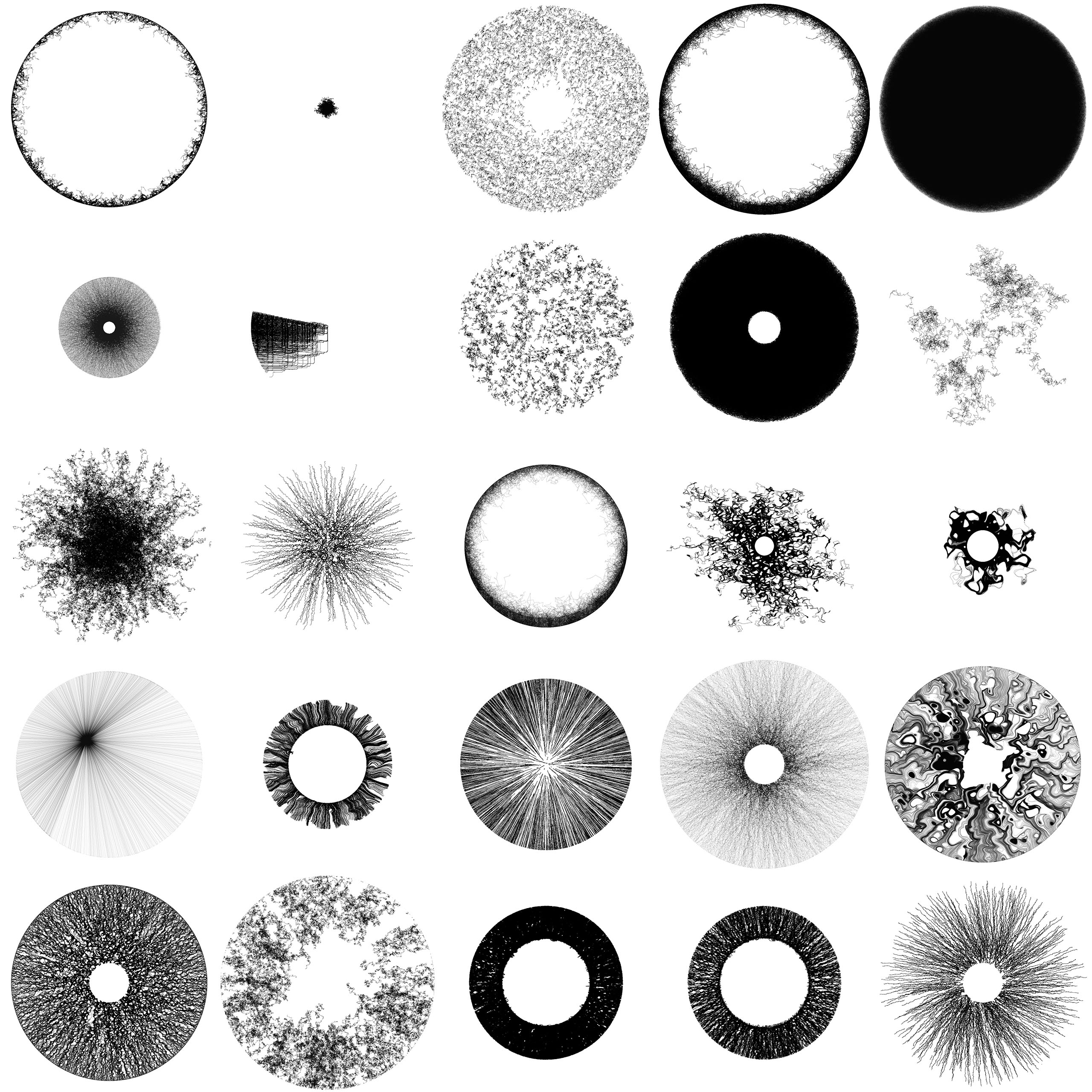}
        \\
        a & b & c \\
    \end{tabular}
    \caption{Snap shots of the elite population at a) generation $e = 1$, b) generation $e = 10$ and c) generation $e = 50$}
    \label{fig:feature-space}
\end{figure*}

Fig. \ref{fig:feature-space} shows the state of the feature space at three different stages of the process: after the initial population is evaluated (a) after $e = 10$ generations (b) and at the end of the evolutionary process (c). The obvious observation is the increase in diversity, represented by the number of classes that are populated on the image on the right. More subtle is the change in the classes that were initially populated, where more complex patterns can be observed.

The results suggest that the proposed approach has potential to be used as a tool for creative exploration of complex design spaces. In order to further understand how capable this system is, we conducted a `control' experiment where diversity is turned-off and the system runs exclusively for fitness, much like a traditional genetic algorithm. Selection was achieved by sorting the generated population by fitness and then sampling from the fittest 25\% of the population. Each sampled individual is then mutated using the same parameters we used for the MAP-Elites runs ($r = 1/g_l$ and $f = 0.25$). A total of 20 evolutionary runs of $e = 50$ generations each, and a population of $\lambda = 15$ individuals were executed using this `GA' approach.

When analysing the characteristics of the fittest drawings obtained through this method, shown in Fig.~\ref{fig:ga-fittest}, it is possible to observe that the fitness function tends to favour features like large size circles and high pixel density, as well as similar textures, yielding designs that look similar to each other, potentially disregarding `good' alternatives that do not exhibit these traits. 

\section{Discussion \& Future Work}
\label{s:discussion}
\begin{figure}
    \centering
    \includegraphics[width=0.95\columnwidth]{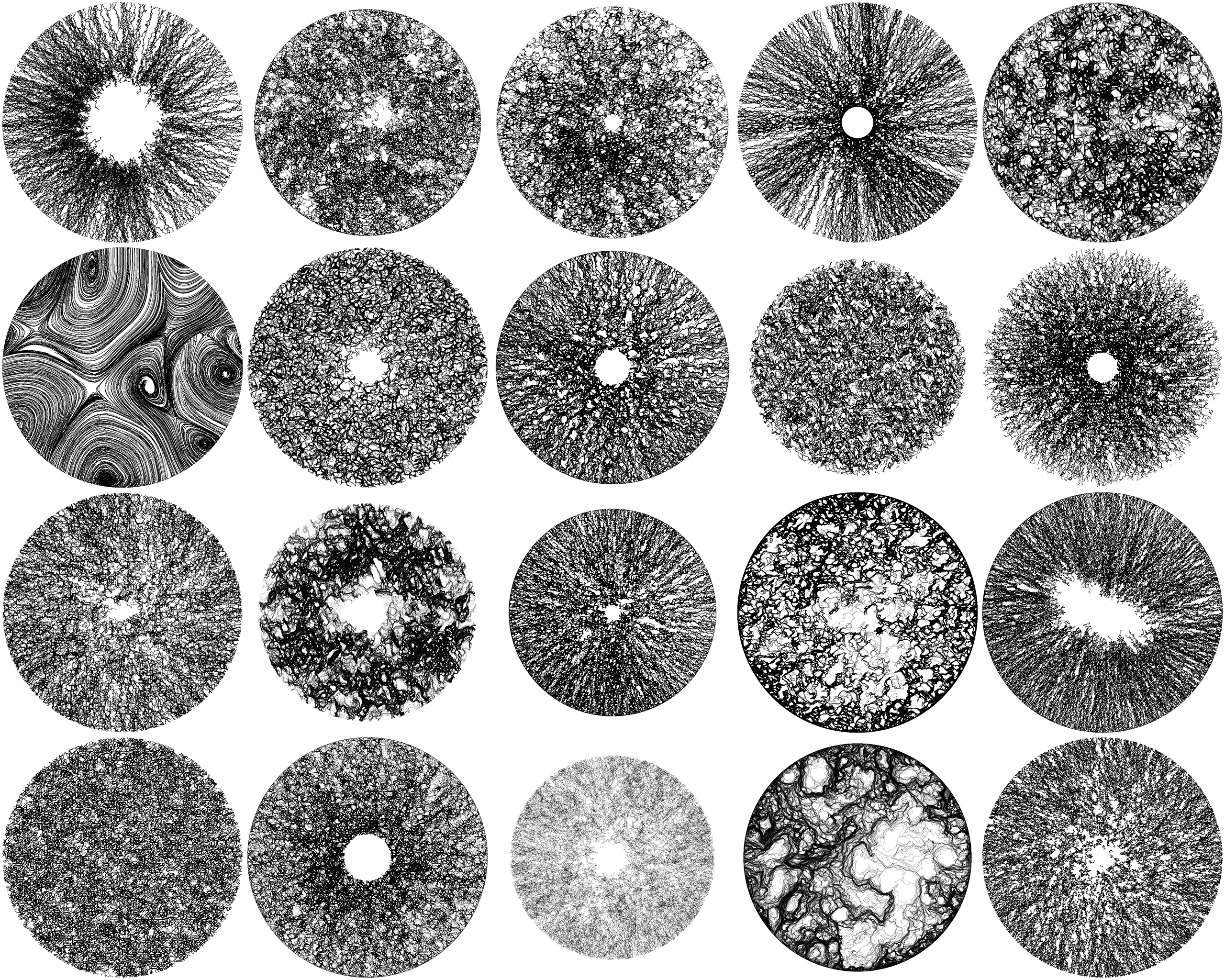}
    \caption{Fittest individual for every fitness-only run}
    \label{fig:ga-fittest}
\end{figure}

While our results are very promising, there are a number of aspects of our current implementation that require further development. Firstly, we use a relatively simple fitness function, which, as shown in Fig. \ref{fig:ga-fittest}, is biased toward a small set of specific designs. A more sophisticated measure of fitness, such as a weighted function combining multiple image complexity metrics, like histograms and skew (see \cite{McCormack2022b} for reference) could  be more effective in identifying a more nuanced range of alternatives as quality candidates. This would enable the system to widen the diversity of the elite population, increasing the probability for the user to encounter unexpected emergent in them.

Secondly, the architecture of the VAE used in this work is relatively simple, which limits its accuracy for feature extraction. By increasing the complexity of the model it would be possible to train a more precise feature extractor, especially for complex datasets.

Finally, to reduce the VAE vectors from 512-D to 2-D  we use a combination of PCA and TSNE. This process led to a loss of detail in the representation of the feature space, which resulted in inaccurate clustering, reflected in the similarity of drawings found in distinct classes (see drawings on the bottom row, fourth column in Fig. \ref{fig:feature-space}c). Further improvements to the system could be made by exploring different dimensionality techniques such as UMAP or different clustering methods like deep variational clustering.

\bibliographystyle{ACM-Reference-Format}
\bibliography{bibliography} 

\end{document}